\documentclass{sig-alternate-05-2015}

\usepackage{tikz}
\usepackage{booktabs}
\usepackage{multirow}

\begin{document}






%

\title{Distributed Deep Learning for Question Answering}
%
%
%
%
%

\numberofauthors{3} 
%
\author{\alignauthor
       Minwei Feng\\
       \affaddr{IBM Watson, Yorktown Heights, NY, USA, 10598}\\
       \email{mfeng@us.ibm.com}
\alignauthor
Bing Xiang\\
       \affaddr{IBM Watson, Yorktown Heights, NY, USA, 10598}\\
       \email{bingxia@us.ibm.com}
\alignauthor 
  Bowen Zhou\\
       \affaddr{IBM Watson, Yorktown Heights, NY, USA, 10598}\\
       \email{zhou@us.ibm.com}
       }
%
%


\maketitle
\begin{abstract}
This paper is an empirical study of the distributed deep learning for question answering subtasks: answer selection and question classification. Comparison studies of SGD, MSGD, ADADELTA, ADAGRAD, ADAM/ADAMAX, RMSPROP, DOWNPOUR and EASGD/EAMSGD algorithms have been presented. Experimental results show that the distributed framework based on the message passing interface can accelerate the convergence speed at a sublinear scale. This paper demonstrates the importance of distributed training. For example, with 48 workers, a 24x speedup is achievable for the answer selection task and running time is decreased from 138.2 hours to 5.81 hours, which will increase the productivity significantly.
\end{abstract}

%
%


%
%

%
%



\section{Introduction}
\label{label:intro}
\footnote{This paper will appear in the Proceeding of The 25th ACM International Conference on Information and Knowledge Management (CIKM 2016),  Indianapolis, USA.}
Deep Learning technology \cite{dp} has been widely adopted in various AI tasks and has achieved the state-of-the-art performance. One practical challenge of Deep Learning is the highly time consuming training procedure. It is not unusual to see the reported training time in the magnitude of days or even weeks in research papers. However, this is rarely acceptable for practical commercial usage (e.g. training as a service on the cloud) where short turn around time is expected by customers. Even for research environment the long time computation could stop scientists from running as many experiments as needed and slow down the R\&D cycle. Hence the distributed training has become a crucial research direction along with the advancement of deep learning itself on the algorithm side.  

Various infrastructures and experimental results have been published recently. Most of those results are on computer vision benchmark tasks like CIFAR10
or ImageNet. In this paper, we focus on the question answering (QA) domain. We study two subtasks of QA: answer selection and question classification. It is trivial to observe the epoch speed (training data processing speed) increased after more computing resources have been adopted. However, this does not necessarily guarantee that the convergence speed is also improved. The ultimate goal is to have convergence speedup as  users will expect models with equal accuracy to be trained faster when the cost is increased for more computing resources. Many optimization algorithms are available but their performances  have not been compared under the distributed training mode. The motivation of this paper is to conduct comparison study for distributed training algorithms and demonstrate the sublinear scalability of the distributed training on convergence speed.
We have compared the latest technologies, including SGD \cite{Bottou:1999}~, MSGD \cite{Hinton_icml2013}~, RMSPROP  \cite{rmsprop}, ADADELTA \cite{adadelta}, ADAGRAD \cite{adagrad}, ADAM/ADAMAX \cite{Adamadamax}, DOWNPOUR \cite{DeanCMCDLMRSTYN12} and EASGD/EAMSGD \cite{ZhangCL14a}. To our best knowledge, it is the first time that such results of distributed training algorithms have been reported on the QA subtasks. 

The rest of the paper is organized as follows: section \ref{label:related_work} is the summary of related work; section \ref{label:task} will describe the answer selection benchmark task; section \ref{label:nlctask} summarizes the question classification task; we demonstrate the MPI-based infrastructure in section \ref{label:mpi} and the review of the distributed training algorithms is given in section \ref{label:algs}
. Experimental results are reported in  section \ref{label:exps} and finally conclusions are drawn in section \ref{label:conc}~.

\pgfdeclarelayer{background}
\pgfdeclarelayer{foreground}
\pgfsetlayers{background,main,foreground}

\tikzstyle{sensor}=[draw, fill=blue!20, text width=5em, 
    text centered, minimum height=2.2em] 
    
\tikzstyle{tester}=[draw, fill=green!40, text width=5em, 
    text centered, minimum height=2.2em]

\tikzstyle{master} = [sensor,  fill=red!20, 
    rounded corners]

\begin{figure*}[t]
\centering
\begin{minipage}[t]{.45\textwidth}
\centering
\begin{tikzpicture}[scale=0.8]
  \begin{scope}
    \draw (0.0,0.0)rectangle (0.4,1.2) node[pos=.5] {\scriptsize A};  
    \draw (0.0,1.8)rectangle (0.4,3.0) node[pos=.5] {\scriptsize Q};   
    \draw (0.8,0.0)rectangle (2.0,3.0) node[pos=.5] {\scriptsize HL$_{QA}$};   
    \draw (2.5,0.0)rectangle (3.7,3.0) node[pos=.5] {\scriptsize CNN$_{QA}$};
    \draw (4.0,0.0)rectangle (4.4,1.2) node[pos=.5] {\scriptsize \begin{tabular}{c} P \\ $+$ \\ R \end{tabular}};
    \draw (4.0,1.8)rectangle (4.4,3.0) node[pos=.5] {\scriptsize \begin{tabular}{c} P \\ $+$ \\ R \end{tabular}};
    
    \draw [->, thick] (0.4,0.6) -- (0.8,0.6);
    \draw [->, thick] (2.0,0.6) -- (2.5,0.6);
    \draw [->, thick] (0.4,2.4) -- (0.8,2.4);
    \draw [->, thick] (2.0,2.4) -- (2.5,2.4);
    \draw [->, thick] (3.7,0.6) -- (4.0,0.6);
    \draw [->, thick] (3.7,2.4) -- (4.0,2.4);
    \draw [->, thick] (4.4,0.6) -- (5.2,1.4);
    \draw [->, thick] (4.4,2.5) -- (5.2,1.8);
    \fill (5.5,1.6) circle[radius=1pt] ;
    \draw (5.5,1.6) circle[radius=10pt, thick];
    \node at (5.7,2.1) {\scriptsize Similarity};
  \end{scope}
\end{tikzpicture}
\caption{\small Architecture for answer selection. HL is the hidden layer with $Tanh$ activation function. CNN is the convolutional neural networks. P stands for maxpooling and R stands for the $ReLU$ activation function. QA means the weights of corresponding layer are shared by Q and A.}
\label{fig:arc}
\end{minipage}\hfill
\begin{minipage}[t]{.45\textwidth}
\centering
\begin{tikzpicture}[scale=0.8]
  \begin{scope}
    \draw (0.0,0.0)rectangle (0.4,3.0) node[pos=.5] {\scriptsize Q};     
    \draw (0.8,0.0)rectangle (1.6,3.0) node[pos=.5] {\scriptsize HL};   
    \draw (2.0,0.0)rectangle (3.0,3.0) node[pos=.5] {\scriptsize CNN};
    \draw (3.5,0.0)rectangle (4.3,3.0) node[pos=.5] {\scriptsize HL};
    \draw (4.8,0.0)rectangle (5.6,3.0) node[pos=.5] {\scriptsize \begin{tabular}{c} Soft\\ max  \end{tabular}};    
    \draw [->, thick] (0.4,1.5) -- (0.8,1.5);
    \draw [->, thick] (1.6,1.5) -- (2.0,1.5);
    \draw [->, thick] (3.0,1.5) -- (3.5,1.5);
    \draw [->, thick] (4.3,1.5) -- (4.8,1.5);
  \end{scope}
\end{tikzpicture}
\caption{\small Architecture for question classification. HL is the hidden layer with $Tanh$ activation function. CNN is the convolutional neural networks.}
\label{fig:arc2}
\end{minipage}
\vspace{-0.3cm}
\end{figure*}

\section{Related Work}
\label{label:related_work}
Various systems have been proposed for distributed deep learning. One of the pioneering work is Google's Distbelief system \cite{DeanCMCDLMRSTYN12} in which DOWNPOUR has been proposed. The system has multiple parameter servers and clients. Most of other work follow the same spirit of DOWNPOUR. The system Adam \cite{Chilimbi} is another similar framework which has many engineering features like reduced memory copies and mitigating the impact of slow machines. 
IBM's Rudra system \cite{2015arXiv150904210G} is a master-client based distributed framework where the servers are organized as a tree structure to save communication overhead. A parameter server framework is proposed in \cite{limu} that supports flexible consistency models, elastic scalability and continuous fault tolerance. \cite{limu} provides the APIs so that other framework like MXNet\footnote{https://github.com/dmlc/mxnet} can utilize it. The platform Petuum \cite{Xing:2015} supports a synchronization model with bounded staleness. Compared to the previous work, the main contribution of this paper is that we study a different task, answer selection, and focus on the comparison of state-of-the-art algorithms.

\section{Answer Selection Task}
\label{label:task}
Different from many previous work, we study a QA task: answer selection. The paper \cite{Feng2015} created an open task (including the released corpus) which serves as a benchmark for comparison purpose. For the detailed description of the data and task please refer to \cite{Feng2015}. A summary is given here to make the paper self-contained.
Given a question $q$ and an answer candidate pool $\{a_1, a_2, ... , a_s\}$ for that question ($s$ is the pool size), the goal is to find the best answer candidate $a_k$, $1 \leq k \leq s $~. If the selected answer $a_k$ is inside the ground truth set of $q$ (questions could have multiple correct answers), the question $q$ is considered correct. In this paper the best architecture (Figure \ref{fig:arc}) from \cite{Feng2015} has been used. The idea is to learn a vector representation of a given question and its answer candidates and then use a similarity metric to measure the matching degree. The similarity metric is Geometric mean of Euclidean and Sigmoid Dot product (GESD) $k(x, y)=\frac{1}{1+\|x-y\|}\cdot\frac{1}{1+exp(-(x^{\intercal}y + 1))}$. $x$ and $y$ are the vector representations of Q and A.  
The training is computational expensive due to the usage of the hinge loss: for each training question {\small$Q$} there is a positive answer {\small$A^+$}(the ground truth). A training instance is then constructed by pairing this {\small$A^+$} with a negative answer {\small$A^-$}(a wrong answer) sampled from the whole answer space. The forward pass calculation generates vector representations for the question and the two candidates: {\small$V_{Q}$}~, {\small$V_{A^+}$} and {\small$V_{A^-}$}~.  The similarities {\small$GESD(V_{Q},V_{A^+})$} and {\small$GESD(V_{Q},V_{A^-})$} are calculated and their difference is compared to a margin $m$: {\small$GESD(V_{Q},V_{A^+}) - GESD(V_{Q},V_{A^-}) < m$}~. If this condition is not satisfied, there is no update to the model and a new negative example is sampled until the margin is less than $m$ ( this repetitive  negative sampling procedure is time-consuming and  to reduce running time we set maximum sampling times to be 100).

\section{Question Classification Task}
\label{label:nlctask}
The second QA subtask we study in this paper is question classification. For certain application scenario (e.g. online customer service), the set of possible answers for all incoming questions is limited and predefined. Hence we can convert the QA into a question classification problem, where each question's label represents the specific answer in the predefined set. Usually there is a \textit{noAnswer} label in the set for chit-chat questions. The data we used for this task is a customer corpus in financial domain. There are 78566 questions and the answer set size is 6763 (6763 different labels for the classifier). We further randomly split the data into train/valid/test parts with the question size 74566/2000/2000. We use a convolutional neural networks based model (Figure \ref{fig:arc2}) to tackle this task. The last layer is \textit{Softmax} since this is a classification task. Please note that this general model can be applied to many natural language classification tasks such as relation classification, intent classification and sentiment analysis.

\begin{figure}[!t]
\vspace{-0.4cm}
\centering
\begin{tikzpicture}
\path (0,-3em) node (c1) [tester] {Tester};
\path (0,0) node (c1) [sensor] {Client 1};
\path (0,3em) node (c2) [sensor] {Client 2};
\path (0,5.6em) node (d1)  {\vdots};
\path (0,8em) node (cn) [sensor] {Client n};
\path (3.5,0) node (s1) [master] {Server 1};
\path (3.5,3em) node (s2) [master] {Server 2};
\path (3.5,5.6em) node (d1)  {\vdots};
\path (3.5,8em) node (sn) [master] {Server n};

\draw [-, thick] (0.9,0) -- (2.6,2.6);
\draw [-, thick] (0.9,0) -- (2.6,1.0);
\draw [-, thick] (0.9,0) -- (2.6,0);
\draw [-, thick] (0.9,1.0) -- (2.6,2.6);
\draw [-, thick] (0.9,1.0) -- (2.6,1.0);
\draw [-, thick] (0.9,1.0) -- (2.6,0);
\draw [-, thick] (0.9,2.6) -- (2.6,2.6);
\draw [-, thick] (0.9,2.6) -- (2.6,1.0);
\draw [-, thick] (0.9,2.6) -- (2.6,0);
\draw [-, thick] (0.9,-1.1) -- (2.6,2.6);
\draw [-, thick] (0.9,-1.1) -- (2.6,1.0);
\draw [-, thick] (0.9,-1.1) -- (2.6,0.0);
\begin{pgfonlayer}{background}
        \path[fill=yellow!20,rounded corners, draw=black!50, dashed]
       (-1.4,-1.8) rectangle (4.9,3.7);
\end{pgfonlayer}
\end{tikzpicture}  
\caption{MPI framework}
\label{fig:mpi}
\vspace{-0.2cm}
\end{figure}
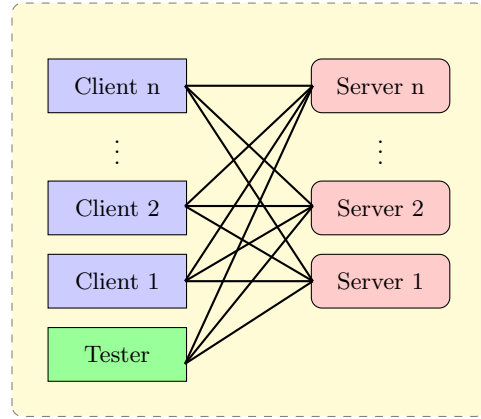

\section{MPI-based Framework}
\label{label:mpi}
Figure \ref{fig:mpi} demonstrates the Message Passing Interface (MPI) framework. 
There are three types of process: worker, parameter server and tester. Those processes are allocated across the high performance computing clusters. The worker will conduct the forward pass/backward pass calculations and send the update messages to servers. The servers hold a central model. They receive the messages from workers and update the central model and send the latest model back to the workers. A tester will only receive the latest model from the servers and run testing over the test corpus periodically. For MPI the  non-blocking communication (\texttt{MPI\char`_ISend}/\texttt{MPI\char`_IRecv}) is used to increase the overall speed.  To reduce the communication overhead,  we split the model into partitions and set up multiple servers. Each server is responsible for the storage and update of one model partition. The amount of worker and server is set to be equal. We use the popular MPI toolkit MPICH. 

\section{Distributed Training Algorithms}
\label{label:algs}
We have compared state-of-the-art algorithms: stochastic gradient descent (SGD) \cite{Bottou:1999}~, momentum stochastic gradient descent (MSGD) \cite{Hinton_icml2013}, RMSPROP (implemented same as section 4.2 of \cite{rmsprop}), ADADELTA \cite{adadelta}, ADAGRAD \cite{adagrad}, ADAM/ADAMAX \cite{Adamadamax}, DOWNPOUR \cite{DeanCMCDLMRSTYN12}~, elastic averaging stochastic gradient descent (EASGD) and its variation momentum EASGD (EAMSGD) \cite{ZhangCL14a}~. 

\begin{table}[t]
\centering
\scriptsize
\begin{tabular}{lrrr}
\toprule
\multirow{2}{*}{\bf Method }  & {\bf Peak } &\multirow{2}{*}{\bf Time (hour)}&
\multirow{2}{*}{\bf Comment} \\
& {\bf Accuracy(\%)} & & \\
\hline \\
SGD          & 64.61  &  145.98  &  61.50@50.95 \\
MSGD         & 65.50  &  138.20  &  61.50@54.07 \\
RMSPROP      & 64.89  &  99.06   &  61.50@37.03 \\
ADADELTA     & 61.50  &  49.83   &    \\
ADAGRAD      & 58.50  &  126.52  &    \\
ADAM         & 54.06  &  140.00  &    \\
ADAMAX       & 52.06  &  145.55  &    \\
\bottomrule
\end{tabular}
\caption{Answer selection task: single worker training results. Accuracy is the peak accuracy of test corpus within 6 days. Time is the wall clock time of the peak accuracy.}
\label{table:single_results}
\end{table}

\begin{table}[t]
\centering
\scriptsize
\begin{tabular}{lrrr}
\toprule
\multirow{2}{*}{\bf Method }  & {\bf Peak } &\multirow{2}{*}{\bf Time (hour)}&
\multirow{2}{*}{\bf Comment} \\
& {\bf Accuracy(\%)} & & \\
 \hline \\
DOWNPOUR   & 64.61     & 5.26    &     \\
EASGD      & 66.11     & 8.57    &  65.50@8.23     \\
EAMSGD     & 67.50     & 11.09   &  65.50@5.81   \\
RMSPROP    & 64.44     & 5.30    &              \\
ADADELTA   & 61.39     & 6.34    &              \\
ADAGRAD    & 59.50     & 8.73    &   58.50@3.75 \\
ADAM       & 55.28     & 9.95    &   54.28@5.82 \\
\bottomrule
\end{tabular}
\caption{Answer selection task: distributed training results with 48 workers. Accuracy is the peak accuracy of test corpus within 12 hours. Time is the wall clock time of the peak accuracy.}
\label{table:distri_results}
\end{table}

\section{Experimental Results}
\label{label:exps}
Table \ref{table:single_results} demonstrates the results of conventional optimization algorithms which use only one worker for the answer selection task.
Table \ref{table:distri_results} demonstrates the results of 
distributed optimization algorithms for the answer selection task. Similarly, the results of the question classification task are shown in Table \ref{table:nlc_single_results} and Table \ref{table:nlc_distri_results}. Each method has its own hyper parameters. We have conducted extensive tuning experiments and only the best results of each method are presented in all tables. The strategy of hyper parameter tuning is two steps of grid search. In the first step, a coarse-grained grid selection of hyper parameters is conducted to find the rough range of the best hyper parameters. Then in the second step, a fine-grained grid selection of hyper parameters is conducted within the range that are discovered in the first step. For the answer selection task, \textbf{Peak Accuracy} is the top accuracy score on the test1 corpus of the released corpus from \cite{Feng2015} within the whole running period. For the question classification task, \textbf{Peak Accuracy} is the top accuracy score on the test corpus within the whole running period.  \textbf{Time} is the wall clock time (unit is hour) when the accuracy reaches that peak value. In \textbf{Comment}, 65.50@8.23 means the accuracy climbs up to 65.50\% at wall clock time 8.23 hours. 
For the answer selection task, we let the single worker training methods keep running for 150 hours (approximately 6 days); for the question classification task, the single worker training methods are set to keep running for 3 days. For distributed methods the running time limit is set to 12 hours for both tasks. This is to save the computing resources so that more experiments can be scheduled. Also in practice it is much less meaningful if the running time is still prohibitive when large amount of computing resources are used.
Finally, from previous study we notice that: for the answer selection task, the highest accuracy scores of test1 corpus are around 65\%; for the question classification task, the model accuracy on test corpus should be around 98.5\%.

\begin{table}[t]
\centering
\scriptsize
\begin{tabular}{lrrr}
\toprule
\multirow{2}{*}{\bf Method }  & {\bf Peak } &\multirow{2}{*}{\bf Time (hour)}&
\multirow{2}{*}{\bf Comment} \\
& {\bf Accuracy(\%)} & & \\
\hline \\
SGD          & 98.60  &  39.02   & 98.50@33.30  \\
MSGD         & 98.50  &  38.30   &   \\
RMSPROP      & 98.70  &  40.97   & 98.50@29.15  \\
ADADELTA     & 93.10  &  70.84   &    \\
ADAGRAD      & 98.50  &  37.57  &    \\
ADAM         & 91.90  &  57.06  &    \\
ADAMAX       & 82.70  &  60.53  &    \\
\bottomrule
\end{tabular}
\caption{Question classification task: single worker training results. Accuracy is the peak accuracy of test corpus within 3 days. Time is the wall clock time of the peak accuracy.}
\label{table:nlc_single_results}
\end{table}

\begin{table}[t]
\centering
\scriptsize
\begin{tabular}{lrrr}
\toprule
\multirow{2}{*}{\bf Method }  & {\bf Peak } &\multirow{2}{*}{\bf Time (hour)}&
\multirow{2}{*}{\bf Comment} \\
& {\bf Accuracy(\%)} & & \\
 \hline \\
DOWNPOUR   & 98.55     & 5.20    &     \\
EASGD      & 97.65     & 10.06   &    \\
EAMSGD     & 98.35     & 4.80    &     \\
RMSPROP    & 98.30     & 4.87    &     \\
ADADELTA   & 97.40     & 5.42    &      \\
ADAGRAD    & 82.65     & 11.61    &   \\
ADAM       & 88.45     & 2.82    &    \\
\bottomrule
\end{tabular}
\caption{Question classification task: distributed training results with 48 workers. Accuracy is the peak accuracy of test corpus within 12 hours. Time is the wall clock time of the peak accuracy.}
\label{table:nlc_distri_results}
\end{table}

\subsection{Results of Answer Selection Task}
In Table \ref{table:single_results}, we observe the following facts from the single worker experiments: (1) 
in terms of peak accuracy, SGD, MSGD and RMSPROP have scores around 65\% which is same with the highest number reported in \cite{Feng2015}; (2) ADADELTA and ADAGRAD lose several points of accuracy; (3) ADAM and ADAMAX perform significantly worse than other methods; (4) if a top accuracy is the goal, the best method is MSGD; (5) if for some practical applications where light accuracy loss is acceptable(e.g. 61.50\% is fine), then RMSPROP is preferable as it converges faster. 

Since ADAMAX does not work well and is similar to ADAM, we did not conduct experiments using distributed versions of ADAMAX algorithm. Also notice the algorithms EASGD/EAMSGD are only designed for the distributed training.  In Table \ref{table:distri_results}, we observe the following facts from the 48-worker experiments: (1) overall distributed training does not incur accuracy loss; (2) EASGD/EAMSGD achieve higher peak accuracy than the best score 65.50\% from single worker results;
(3) overall distributed training speeds up the training;
(4) for EAMSGD it takes 5.81 hours to reach 65.50\% and compared to single worker MSGD where it takes 138.20 hours to climb up to 65.50\% , a 24x speed up is achievable by using distributed training.

\subsection{Results of Question Classification Task}
In Table \ref{table:nlc_single_results}, we observe the following facts from the single worker experiments: (1) 
in terms of peak accuracy, SGD, MSGD, RMSPROP and ADAGRAD have scores around 98.50\% which is the expected accuracy score from previous study; (2) ADADELTA and ADAM lose several points of accuracy; (3) ADAMAX performs significantly worse than other methods.

In Table \ref{table:nlc_distri_results}, we observe the following facts from the 48-worker experiments: (1) in terms of accuracy DOWNPOUR, EASGD, ADADELTA, EAMSGD and RMSPROP perform well; (2) ADAGRAD performs poorly under the distributed scenario where large accuracy loss is incurred; (3) considering both accuracy and convergence speed, DOWNPOUR, EAMSGD and RMSPROP are outstanding. Overall the training time has been decreased from 29.15 hours to 5.2 hours. The improvement is less compared to the answer selection task but it is still a significant productivity boost in practice.

\section{Conclusions}
\label{label:conc}
We have conducted an empirical study of the distributed training for the answer selection task and question classification task which are crucial components of QA. We build the framework with MPI. The state-of-the-art algorithms have been compared, including SGD, MSGD, RMSPROP, ADADELTA, ADAGRAD, ADAM, ADAMAX, DOWNPOUR and EASGD/EAMSGD. To our best knowledge, it is the first time that the  experimental results for distributed training have been reported on QA subtasks. This work proves the significance of the distributed training and a proper algorithm selection is crucial. E.g., for the answer selection task, a 24x speedup is achievable with the deployment of 48 workers and running time is decreased from 138.2 hours to 5.81 hours which is a huge gain for practical productivity. We realize that due to the lack of a solid mathematical foundation, the distributed training is still a trial-and-error procedure. Our experiences show that the hyper parameter tuning (especially the learning rate) can play a crucial role for the performance. On the other hand, the task itself could change the performance. For example, in \cite{Adamadamax} the ADAM demonstrates superior performance for image classification tasks while in our study the performance of ADAM/ADAMAX is relatively weak. From the four tables we can reach the conclusion that DOWNPOUR, EAMSGD and RMSPROP are the most attractive distributed training methods as they significantly increase the convergence speed while maintain the accuracy. The code in this paper has been written based on the Torch7 framework and our source code will be released. For future work we plan to study an algorithm combination strategy so that different distributed training methods could benefit from each other and further improvement could be achieved.


%
\bibliographystyle{abbrv}
\bibliography{sig-alternate-sample} 
%
%

\end{document}